\documentclass[conference]{IEEEtran}
\IEEEoverridecommandlockouts
\usepackage{cite}
\usepackage{graphicx}
\usepackage{textcomp}
\usepackage{xcolor}
\usepackage{amsfonts}
\usepackage{tikz}
\usepackage{textcomp}
\usepackage{wrapfig}
\usepackage{comment}
\usepackage{tcolorbox}
\usepackage{hyperref,xcolor}
\usepackage{mdwmath}
\usepackage{mdwtab}
\usepackage{eqparbox}
\usepackage{amsmath,amssymb,amsfonts}
\usepackage{graphicx}
\usepackage{textcomp}
\usepackage{xcolor}
\usepackage{caption}
\usepackage{array}
\usepackage[tight,footnotesize]{subfigure}
\usepackage{tabularx}
\usepackage{lscape}
\usepackage{mathtools}
\usepackage{booktabs}
\usepackage{filecontents}

\usepackage{times}

\usepackage{soul}
\usepackage{url}
\usepackage[utf8]{inputenc}
\usepackage{booktabs}
\usepackage{tikz}

\usepackage[utf8]{inputenc}
\usepackage{graphicx}
\usepackage{booktabs}
\usepackage{booktabs}
\usepackage{algpseudocode}
\usepackage{algorithm}
 \makeatletter
\newcommand{\linebreakand}{%
  \end{@IEEEauthorhalign}
  \hfill\mbox{}\par
  \mbox{}\hfill\begin{@IEEEauthorhalign}
}
\DeclareMathOperator*{\argmax}{arg\,max}

\def\BibTeX{{\rm B\kern-.05em{\sc i\kern-.025em b}\kern-.08em
    T\kern-.1667em\lower.7ex\hbox{E}\kern-.125emX}}
\begin{document}

\title{APARATE: Adaptive Adversarial Patch for CNN-based Monocular Depth Estimation for Autonomous Navigation\\
%{\footnotesize \textsuperscript{*}Note: Sub-titles are not captured in Xplore and
%should not be used}
%\thanks{Identify applicable funding agency here. If none, delete this.}
}

\author{
\IEEEauthorblockN{1\textsuperscript{st} Amira Guesmi}
\IEEEauthorblockA{\textit{eBrain Lab, Division of Engineering} \\
\textit{New York University (NYU) Abu Dhabi}\\
UAE \\
%ag9321@nyu.edu
}
\and
\IEEEauthorblockN{2\textsuperscript{nd} Muhammad Abdullah Hanif}
\IEEEauthorblockA{\textit{eBrain Lab, Division of Engineering} \\
\textit{New York University (NYU) Abu Dhabi}\\
UAE \\
%email address or ORCID
}
\and
\IEEEauthorblockN{3\textsuperscript{rd} Ihsen Alouani}
\IEEEauthorblockA{\textit{CSIT} \\
\textit{Queen’s University Belfast}\\
UK\\
%email address or ORCID
}
\linebreakand%
\IEEEauthorblockN{4\textsuperscript{th} Muhammad Shafique}
\IEEEauthorblockA{\textit{eBrain Lab, Division of Engineering} \\
\textit{New York University (NYU) Abu Dhabi}\\
UAE \\
%email address or ORCID
}
%\and
%\IEEEauthorblockN{5\textsuperscript{th} Given Name Surname}
%\IEEEauthorblockA{\textit{dept. name of organization (of Aff.)} \\
%\textit{name of organization (of Aff.)}\\
%City, Country \\
%email address or ORCID}
%\and
%\IEEEauthorblockN{6\textsuperscript{th} Given Name Surname}
%\IEEEauthorblockA{\textit{dept. name of organization (of Aff.)} \\
%\textit{name of organization (of Aff.)}\\
%City, Country \\
%email address or ORCID}
}

\maketitle

\begin{abstract}
In recent times, monocular depth estimation (MDE) has experienced significant advancements in performance, largely attributed to the integration of convolutional neural networks (CNNs). Nevertheless, the susceptibility of CNNs to adversarial attacks has emerged as a noteworthy concern, especially in domains where safety and security are paramount. This concern holds particular weight for MDE due to its critical role in applications like autonomous driving and robotic navigation, where accurate scene understanding is pivotal. The perturbing effects of adversarial attacks on MDE have prompted researchers to develop strategies involving adversarial patches to assess the vulnerabilities of CNN-based depth prediction methods.
However, the existing approaches fall short of inducing a comprehensive and substantially disruptive impact on the vision system. Instead, their influence is partial and confined to specific local areas. These methods lead to erroneous depth predictions only within the overlapping region with the input image, without considering the characteristics of the target object, such as its size, shape, and position.
This study delves into the susceptibility of MDE to adversarial patches in a more comprehensive manner. We introduce an innovative adaptive adversarial patch named APARATE\footnote{The term "APARATE" is inspired by a teleportation spell in the Harry Potter series, enabling instantaneous travel between locations.}. This patch possesses the ability to selectively undermine MDE in two distinct ways: by distorting the estimated distances or by creating the illusion of an object disappearing from the perspective of the autonomous system. Notably, APARATE is designed to be sensitive to the shape and scale of the target object, and its influence extends beyond immediate proximity.
Our proposed APARATE achieves a mean depth estimation error exceeding $14~meters$, affecting up to $99\%$ of the targeted region. Through this work, we aim to underscore the substantial threat posed by adversarial attacks within the field of MDE. We hope that our findings will raise awareness within the research community about the tangible risks associated with such attacks in real-world scenarios. Furthermore, we anticipate that our study will serve as a catalyst for the exploration and development of more robust and adaptable defense mechanisms for autonomous robotic systems.
\end{abstract}


\section{Introduction}\label{sec:intro}
\begin{figure*}[!ht]
\centering
\includegraphics[width=2\columnwidth]{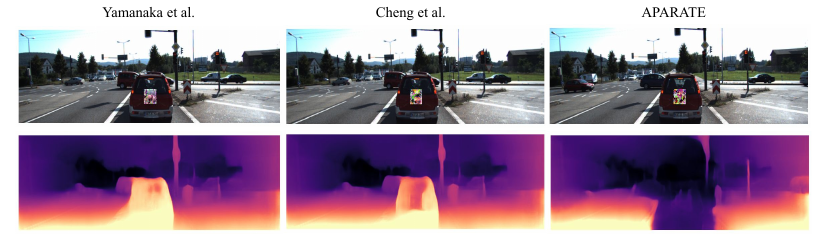}
\caption{Our APARATE makes the object fully disappear, in contrast, adversarial patches proposed by Yamanaka et al. \cite{Yamanaka_Access} and Cheng et al. \cite{Cheng_ECCV} are weak adversarial patches that only impact the depth of a small region of the target object which is restricted to the overlapping region between the patch and the input image.}
\label{results}
\end{figure*}
Monocular depth estimation (MDE) has found increasing utility across various practical applications such as robotics and autonomous driving (AD). MDE involves deriving depth insights from a single image, thereby enhancing scene comprehension. Its significance extends to several critical robotic functions, including obstacle avoidance \cite{obstacles}, object detection \cite{autonomous2}, visual SLAM \cite{robotics1, slam}, and visual relocalization \cite{relocalization}.

Several methodologies for depth estimation rely on technologies like RGB-D cameras, Radar, LiDAR, or ultrasound devices to directly capture depth information within a scene. However, these alternatives exhibit notable shortcomings. RGB-D cameras possess a limited measurement range, LiDAR and Radar deliver sparse data, and both are costly sensing solutions that might not be viable for compact autonomous systems. Ultrasound devices, on the other hand, are marred by inherent measurement inaccuracies. Moreover, these technologies demand substantial energy consumption and feature large form factors, rendering them unsuitable for resource-restricted, small-scale systems that must adhere to stringent real-world design constraints. In contrast, RGB cameras stand out as lightweight and cost-effective options. Importantly, they have the capacity to furnish more comprehensive environmental data.

Prominent players within the autonomous vehicle sector are actively pushing the envelope of self-driving technology by harnessing cost-effective camera solutions. Notably, monocular depth estimation (MDE) has been seamlessly integrated into Tesla's production-grade Autopilot system \cite{tesla2, tesla}. Evidently, other major autonomous driving (AD) enterprises, including Toyota \cite{toyota} and Huawei \cite{huawei}, are following Tesla's footsteps to propel self-driving advancements through this approach.

The surge in the adoption of MDE can be attributed to the widespread utilization of deep learning networks. However, as these networks have demonstrated vulnerabilities to adversarial attacks, safeguarding the security of MDE models becomes an imperative consideration.

Adversarial attacks targeting deep learning systems can be classified into two distinct scenarios: (1) \textit{digital attacks} \cite{fgsm,C&W}, wherein an attacker subtly introduces imperceptible adversarial noise into a digital input image; and (2) \textit{physical attacks} \cite{phy9,phy10,Guesmi_2024_CVPR, advart, saam, Radar}, where the attacker generates an adversarial patch, prints it, and strategically places it within the scene. This manipulated scene is then captured and fed into the victim model through an image. In the scope of this study, we concentrate on the second category, primarily due to its practical applicability in real-world contexts, such as targeting the vision systems of autonomous vehicles or navigation systems in robots.

In physical adversarial examples, the intention is to devise alterations that can be printed out and subsequently re-captured by a camera. Consequently, the attacker's influence is confined to a subset of the pixels that will ultimately reach the victim model. It's noteworthy that the attacker lacks control over factors like perspective, scale, and the image processing steps that cameras inherently execute. This introduces additional complexity since the crafted adversarial examples must exhibit resilience against a myriad of transformations to be efficacious within a physical setting. The transition from digital to physical attacks introduces heightened intricacy, demanding the perturbations to withstand real-world distortions arising from varying viewing distances, angles, lighting conditions, and the limitations of the camera.

These challenges in the physical domain encompass the inherent environmental conditions, encompassing factors such as camera distance and angle. Compounding this, fabrication errors emerge, where all perturbation values must correspond to valid colors reproducible in the tangible world. Furthermore, even if a fabrication device, like a printer, can generate specific colors, there inevitably exists some degree of reproduction error.

Previous efforts for physical-world adversarial attacks \cite{Yamanaka_Access, Cheng_ECCV} yielded relatively feeble adversarial patches, which exhibited a limited impact on the depth estimation of a small portion within a targeted object (e.g., vehicles and pedestrians). The sphere of influence was confined to the area of overlap between the patch and the input image, offering substantial room for enhancement.

Our approach introduces a novel technique for crafting adversarial patches tailored to CNN-based monocular depth estimation. In essence, we generate adversarial patches designed to deceive the target methods, prompting them to erroneously estimate the depth of a designated object (e.g., vehicles, pedestrians), or even to entirely conceal the presence of that object.

\textbf{To summarize, the \textit{distinctive contributions} presented in this study encompass:}
\begin{itemize}
\item We introduce a novel penalized loss function that enhances the efficacy of the patch and expands its impact region (refer to Figure \ref{results}).
\item Our devised attack methodology is generic, making it applicable to various object categories present on public roads. While this paper concentrates on two representative object types—cars and pedestrians—for targeting purposes.
\item The optimization process incorporates an automated patch placement mechanism, as opposed to random placement within the scene. This ensures that the patch is not trained on irrelevant objects.
\item Our proposed patch attains a remarkable mean depth estimation error exceeding 14 meters, effectively influencing up to $99\%$ of the targeted region.
\end{itemize}

An overview of our framework is depicted in Figure \ref{approach}, while a comprehensive description can be found in Section \ref{overview}.   

\section{Proposed Approach}\label{sec:method}
\subsection{Problem formulation}
In monocular depth estimation, when presented with a benign image denoted as $I$, the objective of an adversarial attack lies in causing the depth estimation process to inaccurately predict the depth of the intended object by employing a strategically designed image represented as $I^*$. From a technical standpoint, the adversarial example incorporating the generated patch can be mathematically formulated as follows:

\begin{equation}
    I^* = (1 - M_P) \odot I + M_P \odot P
\end{equation}
$\odot$ is the component-wise multiplication, $P$ is the adversarial patch, and $M_P$ is the patch mask, used to constrict the size, shape, and location of the adversarial patch. 
The adversarial depth, i.e., the output of the victim model $F$ when taking as input the adversarial example is:
\begin{equation}
   %d_{adv} = F_{i,j}(I + R_a T_\theta (P)) 
   d_{adv} = F((1 - M_P) \odot I + M_P \odot P)
\end{equation}
The problem of generating an adversarial example can be formulated as a constrained optimization \ref{eq:adv}, given an original input image $I$ and a MDE model $ F(.) $,:
\begin{equation}
\label{eq:adv}
    arg\min_{P} \left\|P\right\|_p  \\
    s.t. F((1 - M_P) \odot I + M_P \odot P) \neq F(I)\\
    %d_{adv} \neq d
\end{equation}

The goal is to identify a minimal adversarial noise, denoted as $P$, with the specific property that when applied to any object within a designated input domain $U$, it strategically compromises the underlying DNN-based MDE model $F(.)$. This compromise can take the form of either distorting the estimated distance or causing the object to vanish from the prediction.

It's important to note that an analytical solution isn't feasible for this optimization task due to the non-convex nature of the DNN-based model $F(.)$ involved. As a result, the formulation of \ref{eq:adv} can be expressed as follows, allowing for the utilization of empirical approximation techniques to numerically address the problem:
\begin{equation}
\label{eq:formulation}
    \argmax_{P} \sum_{I \in \mathcal{U}} l(F((1 - M_P) \odot I + M_P \odot P), F(I))
\end{equation}
Here, $l$ represents a predefined loss function, and $\mathcal{U} \subset U$ denotes the attacker's training dataset. To tackle this challenge, established optimization methods such as Adam \cite{adam} can be employed to address the problem. During each iteration of the training process, the optimizer iteratively updates the adversarial patch $P$.

\subsection{Overview}
\label{overview}
To facilitate a meaningful comparison, we endeavor to replicate the methodology presented in \cite{Yamanaka_Access}, wherein an adversarial patch was trained for arbitrary placement within the scene. In contrast, the approach proposed by \cite{Cheng_ECCV} involves training a patch designed for consistent placement on the same object at a fixed distance from the camera (not considering the stealthiness constraint, as it falls outside the scope of our work). As clearly illustrated in Figure \ref{results}, our APARATE patch remarkably achieves the complete disappearance of the targeted object. Conversely, we observe a limited impact with the other two patches.
In conducting this comparison, we maintain uniformity by employing identical optimization parameters, patch dimensions, and shapes.

The objective of our study is to create physical adversarial patches possessing a comprehensive impact that encompasses the entirety of the object, irrespective of its dimensions, contours, or placement, while upholding their effectiveness as attack tools. To accomplish this, we introduce the utilization of a pre-trained object detector to identify the precise location of the targeted object—meaning the object we intend to conceal or manipulate its predicted depth. This identification enables the generation of a patch designed to withstand variations in camera-object distance.

Furthermore, we introduce a novel loss function designed to amplify the patch's effect and expand the area it influences.

\begin{figure*}[!ht]
\centering
\includegraphics[width=2\columnwidth]{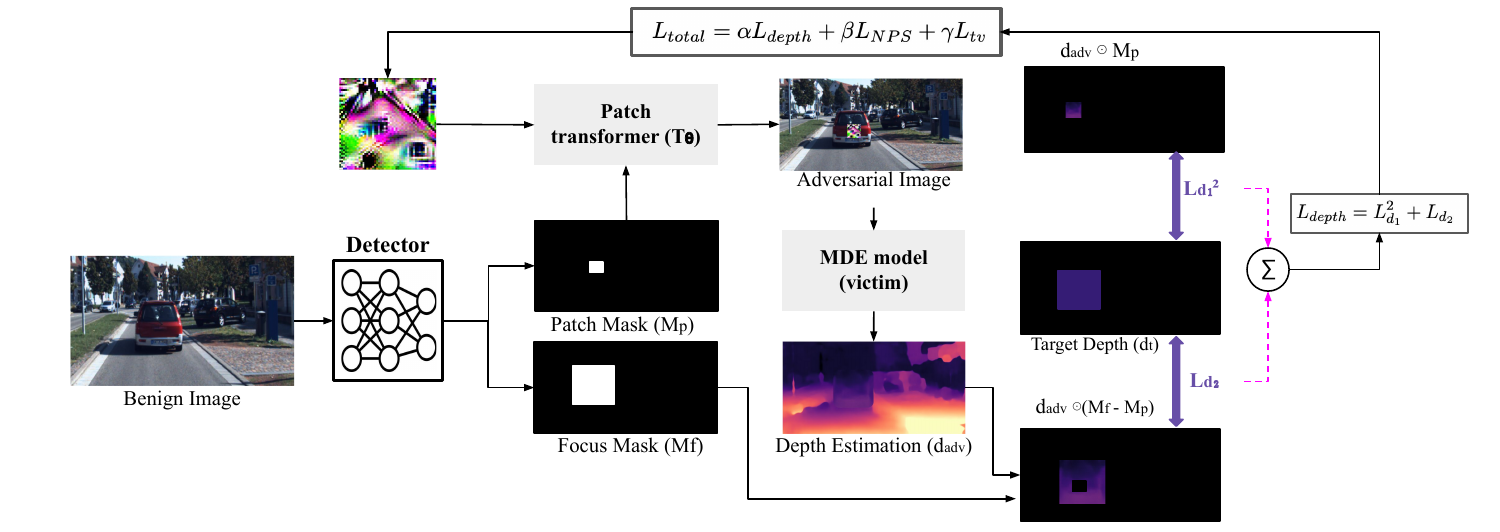}
\caption{\textbf{Overview of the proposed approach:} Given a pre-trained object detector we generate two masks: the patch mask ($M_p$) corresponding to the location of the patch at the center of the target object and the focus mask ($M_f$) corresponding to the object covered area and the attacked region. We feed the patch to the patch transformer and perform the geometric transformations described in Section \ref{GT}. We, later on, render the patch on top of the input image by harnessing information from the object detector as described in section \ref{PA}. After that, we perform a forward pass, i.e., we feed the resulting adversarial image to the MDE model. The next step is to apply the generated masks to compute the required loss functions. Then, we compute the gradient of the patch and based on this information we update the patch $P$.}
\label{approach}
\end{figure*}

As depicted in Figure \ref{approach}, we begin with a pre-trained object detector and initiate the process by creating two distinct masks. The first mask, denoted as $M_p$, represents the precise placement of the patch at the center of the designated target object. Conversely, the second mask, $M_f$, pertains to the specific region covered by the target object – essentially, the region that the target object influences. Subsequently, the patch is inputted into the patch transformer, where we execute the geometric alterations outlined in Section \ref{GT}.

Once these transformations are applied, we leverage the patch applier to overlay the generated patch onto the input image. This task harnesses information sourced from the object detector, as elucidated in Section \ref{PA}. Following this step, we proceed with a forward pass, directing the resulting adversarial image into the MDE model for analysis.

The subsequent stage entails employing the generated masks to calculate the necessary loss functions. Once these computations are completed, we determine the gradient of the patch. This gradient information guides us in updating the actual patch denoted as $P$.

\subsection{Patch Applier}
\label{PA}

In the context of a physical attack scenario, our control over the perspective, scale, and positioning of the adversarial patch concerning the camera is restricted. As a countermeasure, we strive to bolster the resilience of our patch to a wide array of potential scenarios. During the patch training phase, we employ a methodology where the patch is superimposed onto the target object's surface (e.g., the rear of a vehicle or human attire). This approach allows us to simulate diverse scenes characterized by various settings.

The training process encompasses an assortment of transformations, including rotations and occlusions, meticulously integrated to emulate the plausible appearance of our adversarial patch $P$ in a realistic context. Subsequently, leveraging the capabilities of the object detector, we acquire precise object locations (such as vehicles or individuals) within a given image, denoted as $I$. Placing our adversarial patch $P$ onto the identified object becomes feasible at this stage.

Two distinct masks emerge from this process: $M_f$, encircling the object to demarcate its influence, and $M_p$, designed to constrain the patch's attributes, encompassing its location, dimensions, and shape. It is crucial to emphasize that the object detector's involvement pertains solely to the optimization procedure of the patch and is not extended to the actual attack phase.

To eliminate the necessity for manual patch placement onto objects, akin to the approach in \cite{Cheng_ECCV}, we make use of the YOLOv4-tiny object detector pretrained on the MSCOCO dataset \cite{Lin2014MicrosoftCC}. This detector's capabilities streamline the process by automatically identifying object placements, contributing to a more efficient and effective workflow.

Let $U = \{I_i\}_{i=1}^M$ and $V = \{J_j\}_{j=1}^N$ respectively be the M training and N testing images for a particular scene of attack. We run the YOLOv4-tiny object detector on $U$ and $V$ with an objectness threshold of $0.5$ and non-max suppression IoU threshold of $0.4$. This yields the tuples:
\begin{equation}
    T_i^U = \{(B_{i,k}^U)\}_{k=1}^{D_i} , T_j^V = \{(B_{j,l}^V)\}_{l=1}^{E_j}
\end{equation}

for each $I_i$ and $J_j$ , where $D_i$ is the number of detections in $I_i$ (fixed to 14 in our experiment same as in \cite{naturalistic}). and $B_{i,k}^U$ is the bounding box of the $k-th$ detection in $I_i$. % $s_{i,k}^U$ is the objectness score of the $k-th$ detection in $I_i$
(Same for $E_j$ and $B_{j,l}^V$ for $J_j$). % Note that by design $s_{i,k}^U  \ge 0.5$ and $s_{j,l}^V \ge 0.5$.
The sets of all detected objects are:
\begin{equation}
    T^U = \{T_i^U \}_{i=1}^M , T^V = \{T_j^V \}_{j=1}^N
\end{equation}

The information used to optimize P for a scene of attack are the training images $U$ and annotations $T^U$. The patch P is randomly initialized. Given the current P, the patch is rendered on top of each detected object of the chosen class for each training image $I_i$ using the mask $M_{p_{i,k}}^U$. 
$M_{p_{i,k}}^U$ is a matrix of zeros except for the patch location (The center of the patch is the center of the bounding boxes).

The focus mask $M_{f_{i,k}}^U$ is defined as the space limited by the generated bounding boxes $B_{i,k}^U$ in a way that it covers the whole detected object. $M_{f_{i,k}}^U$ is a matrix of zeros except for the targeted regions (objects covered area) where the pixel values are ones. We multiply the generated masks with the adversarial depth map $d_{adv}$ and use the resulting maps to compute the two losses.

\subsection{Patch Transformer}
\label{GT}

The positioning and perspective of a camera on an autonomous vehicle in relation to another vehicle or target object are characterized by continuous variation. The images captured and supplied to the victim model are taken from diverse distances, angles, and lighting conditions. Consequently, any modification introduced by an attacker, such as an adversarial patch, must be resilient to these evolving circumstances. To emulate this, a range of physical transformations is applied, each reflecting different conditions that can arise. These transformations encompass aspects like introducing noise, applying random rotations, altering scales, and simulating variations in lighting. The patch transformer is employed to execute these transformations effectively.

The transformations executed include:
Random Scaling: The patch's dimensions are randomly adjusted to approximately match its real-world proportions within the scene. Random Rotations: The patch $P$ is subjected to random rotations (up to $\pm20^\circ$) centered around the bounding boxes $B_{i,k}^U$. This emulates uncertainties related to patch placement and sizing during printing. Color Space Transformations: Pixel intensity values are manipulated through various color space transformations. These include introducing random noise (within $\pm0.1$ range), applying random contrast adjustments (within the range $[0.8, 1.2]$), and introducing random brightness adjustments (within $\pm0.1$ range).

The resultant image $T_\theta (I_i)$, which undergoes this composite set of transformations, is then passed through the MDE for further analysis and processing.

%==========================
\subsection{Penalized Depth Loss}
%==========================
\label{newloss}

The bounding boxes $B_{i,k}^U$ generated serve as the foundation for creating a focus mask $M_f$, which encompasses the specific region where we intend to modify the predicted depth.

Our objective is to extend the scope influenced by the patch, going beyond mere pixel overlap. To achieve this, we decompose the depth loss $L_{depth}$ into two distinct terms: $L_{d_1}$ and $L_{d_2}$. $L_{d_1}$ represents the loss incurred by pixels that are overlapped by the patch, while $L_{d_2}$ pertains to the loss stemming from pixels that don't overlap.

In order to direct the optimization process towards prioritizing the reduction of the non-overlapped pixel loss $L_{d_2}$, we employ a squaring operation on the term denoting the disparity between the output depth and the target depth, denoted as $\left | d_t - d_{adv} \right | \odot M_P$. This utilization of quadratic functions is strategic, as they exhibit a slower rate of increase (slope or rate of change), consequently delaying the convergence of overlapped pixel loss in comparison to non-overlapped pixels.

The adversarial losses are defined as the distance between the predicted depth and the target depth and calculated as follows:
\begin{equation}
    %\sum_{i,j}\frac{1}{m \times n} 
    L_{d_1} = \left | d_t - d_{adv} \right | \odot M_P 
\end{equation}

\begin{equation}
    %\sum_{i,j}\frac{1}{m \times n} 
    L_{d_2} = \left | d_t - d_{adv} \right |\odot (M_f - M_P)
\end{equation}

\begin{equation}
    %\sum_{i,j}\frac{1}{m \times n} 
    L_{depth} =  L_{d_1}^2 + L_{d_2} 
\end{equation}

%=======================
\subsection{Adversarial Patch Generation}
%======================
We iteratively perform gradient updates on the adversarial patch $(P)$ in the pixel space in a way that optimizes our objective function defined as follows:

\begin{equation}
    L_{total} = \alpha L_{depth} + \beta L_{NPS} + \gamma L_{tv}
\end{equation}

$L_{depth}$ is the adversarial depth loss.
$L_{NPS}$ representing the non-printability score. The $L_{NPS}$ term encourages colors in $P$ to be as close as possible to colors that can be reproduced by a printing device.

It is defined as:
\begin{equation}
    L_{NPS} = \sum_{i,j} min_{c \in C} (\parallel P_{i,j} - c \parallel_2) 
\end{equation}
where $P_{i,j}$ is the pixel (RGB vector) at $(i, j)$ in P, and $c$ is a color vector from the set of printable colors \cite{Thys_CVF}.

where $m$ and $n$ are the image dimensions. 
$L_{tv}$ is the total variation loss on the generated image to encourage smoothness.
It is defined as:
\begin{equation}
    L_{tv} = \sum_{i,j} \sqrt{(P_{i+1,j} - P_{i,j})^2 + (P_{i,j+1} - P_{i,j})^2}
\end{equation}

where the sub-indices $i$ and $j$ refer to the pixel coordinate of the patch $P$.

$\alpha$, $\beta$, and $\gamma$ are hyper-parameters used to scale the three losses. For our experiments, we set $\alpha =1 $, $\beta = 1.5$, and $\gamma = 0.1$.

We optimize the total loss using Adam \cite{adam} optimizer. We try to minimize the object function $L_{total}$ and optimize the adversarial patch. We freeze all weights and biases in the depth estimator and only update the pixel values of the adversarial patch. The patch is randomly initialized. Algorithm \ref{algo1} illustrates the different stages described in the above sections in order to train our adversarial patch.

\begin{algorithm}[!t]
\caption{APARATE.}
\label{algo1}
\begin{algorithmic}[1]
\State \textbf{Input:} Dataset: $U$, MDE model: $F$, number of epochs: $K$, clean input: $I$, baseline depth $d$, adversarial depth $d_{adv}$, object detector: $D$

\State \textbf{Output:} $P$ adversarial patch.

\State Initialize $P \leftarrow random(0,1)$

\For{ $epoch = 0:K$}
    \For{ I in U}
        \State $d = F_{i,j}(I)$
        \State $(M_P,M_f) \leftarrow  D(data)$
        \State $d_{adv} = F_{i,j}((1 - M_P) \odot I + M_P \odot T_\theta (P))$
        \State $L_{d_1} = \left | d_t - d_{adv} \right | \odot M_P $
        \State $L_{d_2} = \left | d_t - d_{adv} \right |\odot (M_f - M_P)$
        \State $L_{depth} =  L_{d_1}^2 + L_{d_2} $
        \State $L_{NPS} = \sum_{i,j} min_{c \in C} (\parallel P_{i,j} - c \parallel_2) $
        \State $L_{tv} = \sum_{i,j} \sqrt{(P_{i+1,j} - P_{i,j})^2 + (P_{i,j+1} - P_{i,j})^2}$
        \State $L_{total} = \alpha L_{depth} + \beta L_{NPS} + \gamma L_{tv}$
        \State compute loss backward
        \State perform optimizer step
        \State set the gradient of the optimizer to 0
        \State clamp(P, [0,1])

\EndFor
\EndFor
\end{algorithmic}
\end{algorithm}

\section{Experimental Results}\label{sec:exp}
%=====================
\subsection{Experimental Setup}
%=====================
In our experimental setup, we employ three distinct Monocular Depth Estimation (MDE) models: namely, monodepth2 \cite{monodepth2}, Depthhints \cite{Depthhints}, and Manydepth \cite{Manydepth}. These models were chosen based on their practicality and the availability of open-source code, the same models featured in the work presented in \cite{Cheng_ECCV}.

All three models adhere to a general U-Net architecture, characterized by an encoder-decoder network design. This architecture enables the incorporation of both high-level abstract features and local information. The encoder component utilizes the pretrained ResNet18 \cite{resnet} model, which was pretrained on the ImageNet dataset \cite{imagenet}. On the other hand, the decoder encompasses multiple convolutional and upsampling layers, with skip connections facilitating the reconstruction of output back to the input's original resolution.

For our evaluations and experiments, we employ real-world driving scenes extracted from the KITTI 2015 dataset \cite{kitti}. This dataset comprises synchronized stereo video recordings alongside LiDAR measurements, all captured from a moving vehicle navigating urban surroundings. The dataset encapsulates an extensive array of road types, encompassing local and rural roads as well as highways. Within these scenes, a diverse range of objects is present, such as vehicles, pedestrians, and traffic lights, allowing for a comprehensive evaluation of the attack performance.

%=====================
\subsection{Evaluation Metrics}
%=====================
To assess the efficacy of our proposed attack, we utilize two distinct metrics: the \textit{mean depth estimation error} ($E_d$) attributed to the target object, and the \textit{ratio of the affected region} ($R_a$). To calculate these metrics, we establish the depth prediction of an unaltered object as a reference point, contrasting it with the depth prediction of the object subject to our adversarial manipulation.

The \textit{mean depth estimation error} signifies the extent to which our proposed patch disrupts the depth estimation accuracy. A larger value in this metric indicates a more effective attack. Similarly, for the \textit{ratio of the affected region}, a higher value corresponds to enhanced attack performance. Notably, we modify the metrics employed in \cite{Cheng_ECCV} to tailor them to our specific experimental context, as the original metrics are not directly applicable.

In contrast to the approach presented in \cite{Cheng_ECCV}, where a patch is trained on a single object situated at a fixed distance from the camera while varying only the background, our strategy involves training the patch for diverse distances. This encompasses objects positioned at various locations relative to the camera, such as close, far, left, right, or center.

To distinguish objects situated closer to the camera, we introduce a threshold. Initially, the output of the disparity map is normalized within the range of $0$ to $1$, where $0$ signifies the farthest point in the image and $1$ represents the closest point. Objects exhibiting a normalized disparity exceeding $0.5$ are categorized as "\textit{close}," while those surpassing this threshold are deemed "\textit{far}." This categorization facilitates the differentiation between objects based on their proximity to the camera.

\textit{The depth estimation error} was measured using the following metrics:
For close objects:
\begin{equation}
    E_{d_c} = \frac{\sum_{i,j}(\arrowvert d - d_{adv} \arrowvert \odot M_f)[d>0.5] }{\sum_{i,j}{M_f[d>0.5] }}
\end{equation}

For far objects:
\begin{equation}
    E_{d_f} = \frac{\sum_{i,j}(\arrowvert d - d_{adv} \arrowvert \odot M_f)[d \leq 0.5] }{\sum_{i,j}{M_f[d \leq 0.5] }}
\end{equation}

The \textit{ratio of the affected region} is gauged by assessing the proportion of pixels whose depth values have experienced alterations exceeding a specific threshold. This assessment is conducted relative to the number of pixels encompassed by the focus mask ($M_f$). The choice of threshold depends on whether the object is categorized as "close" or "far."

For objects classified as "close," any alteration in a pixel's depth value surpassing $0.1$ results in that pixel being deemed as affected. On the other hand, for objects identified as "far," the depth change must exceed $0.01$ for a pixel to be considered affected. To formalize this, we introduce the indicator function $\textbf{I(x)}$, which equals 1 only when the condition $x$ holds true.

Within this context, we differentiate two distinctive terms for analysis.
For "close" objects
\begin{equation}
    R_{a_c} = \frac{\sum_{i,j}\textbf{I}((\arrowvert d - d_{adv} \arrowvert \odot M_f)[d>0.5] > 0.1) }{\sum_{i,j}{M_f[d>0.5] }}
\end{equation}

for "far" objects:
\begin{equation}
    R_{a_f} = \frac{\sum_{i,j}\textbf{I}((\arrowvert d - d_{adv} \arrowvert \odot M_f)[d \leq 0.5] > 0.01 ) }{\sum_{i,j}{M_f[d \leq 0.5] }}
\end{equation}
%=====================
\subsection{Evaluation Results}
%=====================
To evaluate the influence of the penalized depth loss, we employ the monodepth2 model as the target Monocular Depth Estimation (MDE) model. Our patch optimization process is executed over the course of $4000$ epochs, utilizing the Adam optimizer with a learning rate set at $0.01$. The patch is scaled to a factor of $0.2$ during this process.

\begin{figure}[!ht]
\centering
\includegraphics[width=1\columnwidth]{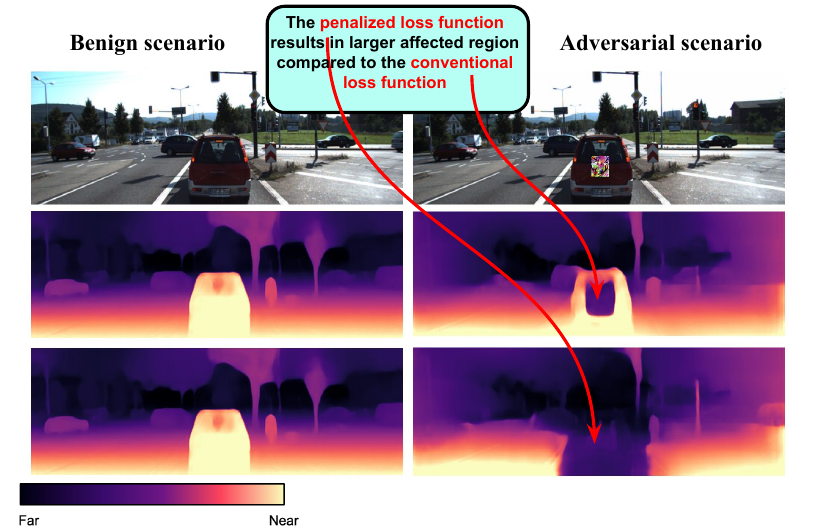}
\caption{Depth prediction w/o the penalized loss function: (Top) the input images, (Middle) results without the penalized depth loss (using the conventional loss), (Bottom) results with the penalized depth loss.}
\label{results_without}
\end{figure}

Initially, a patch is generated through optimization using the conventional loss function. As depicted in Figure \ref{results_without}, the region affected by this patch is confined to the immediate vicinity of the patch itself. This outcome demonstrates a slightly improved performance compared to the patches featured in \cite{Yamanaka_Access, Cheng_ECCV}.

Subsequently, we proceed to assess the effects of the modified depth loss outlined in Section \ref{newloss}. Upon applying this modified depth loss, the resultant patch effectively conceals the entire object, as confirmed by our experimentation.

We extend our attack evaluation to include the three Monocular Depth Estimation (MDE) models, targeting both classes of objects for each model. To commence, we generate adversarial patches specifically tailored for pedestrians and cyclists. In this phase, the patch scale is maintained at $0.2$ for consistent testing across the different models.

\begin{figure}[!ht]
\centering
\includegraphics[width=\columnwidth]{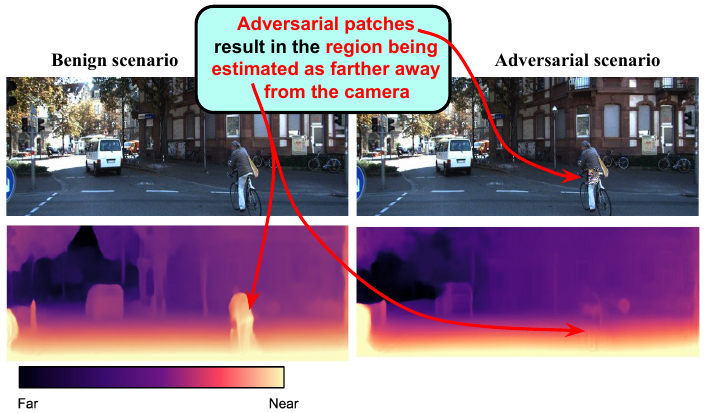}
\caption{Impact of APARATE on pedestrian/cyclist class.}
\label{results_ped}
\end{figure}

Illustrated in Figure \ref{results_ped}, our devised patch demonstrates remarkable effectiveness by entirely concealing the cyclist. Intuitively, smaller objects are comparatively simpler to obscure or manipulate in terms of their depth. However, achieving this for smaller objects necessitates smaller patches, consequently yielding a relatively diminished impact.

In the subsequent experiment, we proceed to craft an adversarial patch targeting the "car" class. Employing the patch scale of $0.2$, we observe that nearly all objects integrated with the APARATE patch achieve complete concealment. This outcome holds true irrespective of the object's specific characteristics, encompassing factors such as shape, size, and proximity to the camera. The extensive success in concealing various objects substantiates the robust nature of our proposed patch.
\begin{figure}[!ht]
\centering
\includegraphics[width=1\columnwidth]{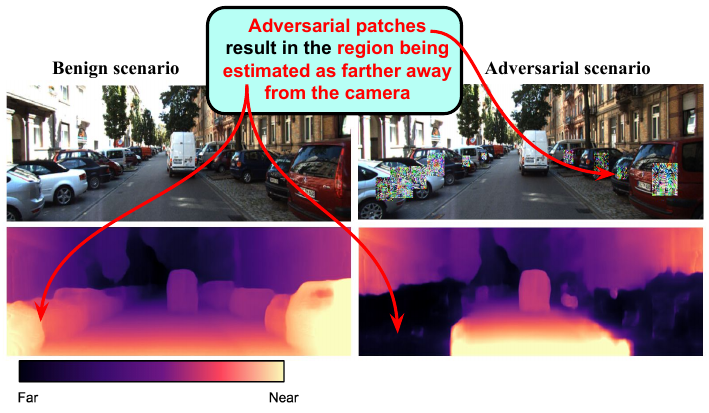}
\caption{Impact of APARATE on car class.}
\label{results_car}
\end{figure}

\subsection{APARATE Performance}
We assess the mean depth estimation errors, denoted as $E_{d_c}$ and $E_{d_f}$, alongside the ratios of the affected regions, designated as $R_{a_c}$ and $R_{a_f}$, for the target object across 100 scenes extracted from the KITTI dataset. The final outcomes are determined by calculating the average values of these metrics, offering a representative result.
\begin{table}[!htp]
  \centering
%\small
  \caption{Attack performance on "close" and "far" objects.}
  \label{performance}
  \begin{tabular}{ccccc}
    \toprule
    \textbf{Models} & \textbf{$E_{d_c}$}  & \textbf{$R_{a_c}$} & \textbf{$E_{d_f}$}  & \textbf{$R_{a_f}$} \\
    \midrule
      \textbf{Monodepth2}  &  0.55 & 0.99  & 0.23  &  0.87\\
      \textbf{Depthhints}  &  0.53 &  0.98 & 0.2  &  0.85\\
      \textbf{Manydepth}   &  0.53  & 0.98  & 0.21  & 0.87\\
  \bottomrule
\end{tabular}
\end{table}

The outcomes we have acquired can be comprehended in the following manner: Hypothetically considering that the camera captures objects as far away as $100$ meters, we categorize objects based on their distance. Those within the range of $[0,50]$ meters, as inferred from our assessment, are labeled as "close", while objects within $[50,100]$ meters are termed "far".

In accordance with the data presented in Table \ref{performance}, when our patch targets a monodepth2-based Monocular Depth Estimation (MDE) model, it achieves an average alteration of approximately $55\%$ for objects within the range of $[0,50]$ meters, and around $23\%$ for objects within $[50,100]$ meters. To illustrate, let's consider an "urban scenario" where a car is traveling at a speed of $40$ km/h. In such a scenario, the braking distance should exceed $24$ meters to ensure the car can come to a safe stop before colliding with another vehicle. However, utilizing our patch, a car positioned $20$ meters away might be incorrectly perceived to be at a distance exceeding $31$ meters, leading to a highly concerning situation of potential collision.

Considering a scenario set on a highway, where the speed limit stands at $160$ km/h, the stipulated safe distance stretches to $96$ meters. When we incorporate our patch, a car positioned $90$ meters away is projected to be situated approximately $110$ meters away. This starkly demonstrates the profound consequences of our devised attack.

Additionally, it's noteworthy that our patch yields an impact on over $98\%$ of the target region for close objects and around $85\%$ for far objects. This data underscores the extensive influence our patch has on altering depth perception.
%--------------------------------
\subsection{APARATE vs Existing Work}
%---------------------------------
We conducted a series of experiments to provide a quantitative comparison between our attack strategy and prior approaches \cite{Cheng_ECCV} and \cite{Yamanaka_Access}. To carry out this comparison, we employed the monodepth2 Monocular Depth Estimation (MDE) model to evaluate both the \textit{depth estimation error} and \textit{ratio of the affected region}. The results presented in Table \ref{performance} demonstrate that our proposed attack consistently achieves the most substantial alteration in depth estimation and encompasses the largest affected region when compared to the referenced prior works.
\begin{table}[!htp]
  \centering
%\small
  \caption{APARATE performance vs. existing attacks.}
  \label{performance}
  \begin{tabular}{ccccc}
    \toprule
    \textbf{Attack} & \textbf{$E_{d_c}$}  & \textbf{$R_{a_c}$} & \textbf{$E_{d_f}$}  & \textbf{$R_{a_f}$} \\
    \midrule
      \textbf{APARATE}         &  \textbf{0.55} & \textbf{0.99} & \textbf{0.23}  &  \textbf{0.87}\\
      \textbf{Cheng et al. \cite{Cheng_ECCV}}     &  0.21 & 0.47 & 0.10  &  0.32\\
      \textbf{Yamanaka et al \cite{Yamanaka_Access}}   &  0.13 & 0.26 & 0.06  & 0.14\\
  \bottomrule
\end{tabular}
\end{table}

Additionally, we employ the Mean Square Error (MSE) as a means to assess the performance of the model in relation to the predicted output depth map derived from an unperturbed input. The formulation of this metric is provided below.
\begin{equation}
    MSE = \frac{1}{n}\sum_{i=1}^{n}(\hat{y_i}-y_i)^2
\end{equation}

As demonstrated in Table \ref{mse}, our attack yields the highest error when juxtaposed with other state-of-the-art attack methods.
\begin{table}[!htp]
\centering
  \caption{MSE of adversarial example compared to the original images for different attacks.}
  \label{mse}
  \begin{tabular}{cccc}
    \toprule
       \textbf{Attack}   & \textbf{APARATE} & \textbf{ \cite{Cheng_ECCV}}  & \textbf{\cite{Yamanaka_Access}}  \\
    \midrule 
           MSE    &  \textbf{0.49}  &  0.12 & 0.05 \\
  \bottomrule
\end{tabular} %}
\end{table}

%-------------------
\subsection{The Influence of Patch Scale}
%--------------------
We subject our attack to evaluation by targeting the "car" object class using three distinct patch sizes: $0.1$, $0.2$, and $0.3$. The mean depth error and the ratio of the affected region are assessed employing the three depth estimation models. Tables \ref{edc} and \ref{edf} showcase the results for mean depth error of "close" and "far" objects, respectively. Notably, an observable trend is that $E_d$ consistently increases alongside the increment in patch size across all three target models. This outcome is in line with expectations, as larger patches exert a more pronounced influence on the resulting depth error. This same trend is echoed in the ratio of affected regions, as reflected in Tables \ref{rac} and \ref{raf}, wherein larger patches correspond to larger affected regions.
\begin{table}[!htp]
  \centering
%\small
  \caption{$E_{d_c}$ for different patch scales.}
  \label{edc}
  \begin{tabular}{cccc}
    \toprule
    \textbf{Scale} & \textbf{Monodepth2}& \textbf{Depthhints}& \textbf{Manydepth}\\
    \midrule
      0.1           & 0.46   & 0.37  & 0.3   \\
      0.2           &  0.55  & 0.53  & 0.53   \\
      0.3           &  0.66 &  0.64 &  0.63\\
  \bottomrule
\end{tabular}
\end{table}

\begin{table}[!htp]
  \centering
  \caption{$R_{a_c}$ for different patch scales.}
  \label{rac}
  \begin{tabular}{cccc}
    \toprule
    \textbf{Scale} & \textbf{Monodepth2}& \textbf{Depthhints}& \textbf{Manydepth}\\
    \midrule
      0.1           &  0.97 &  0.95 & 0.97\\
      0.2           &  0.99  &  0.98 & 0.98 \\
      0.3           &  0.99 & 0.99  & 0.99\\
  \bottomrule
\end{tabular}
\end{table}

\begin{table}[!htp]
  \centering
%\small
  \caption{$E_{d_f}$ for different patch scales.}
  \label{edf}
  \begin{tabular}{cccc}
    \toprule
    \textbf{Scale} & \textbf{Monodepth2}& \textbf{Depthhints}& \textbf{Manydepth}\\
    \midrule
      0.1           &  0.14  & 0.11  & 0.09\\ 
      0.2           &  0.23  & 0.2   & 0.21\\
      0.3           &  0.34  & 0.31  & 0.32\\
  \bottomrule
\end{tabular}
\end{table}

\begin{table}[!htp]
  \centering
  \caption{$R_{a_f}$ for different patch scales.}
  \label{raf}
  \begin{tabular}{cccc}
    \toprule
    \textbf{Scale} & \textbf{Monodepth2}& \textbf{Depthhints}& \textbf{Manydepth}\\
    \midrule
      0.1           & 0.84  & 0.82  & 0.82\\
      0.2           & 0.87  & 0.85  & 0.87\\
      0.3           & 0.89  & 0.87  & 0.88\\
  \bottomrule
\end{tabular}
\end{table}
\section{Discussion}\label{sec:discussion}
To assess the resilience of our patch against existing defense mechanisms, we employ three readily available defense techniques that involve input transformations, without necessitating the re-training of the victim models. Specifically, we apply JPEG compression \cite{jpeg}, introduce Gaussian noise \cite{gaussian}, and utilize median blurring \cite{median}.

In our analysis, we compute and report the mean depth error for benign scenarios, representing cases where the defense technique is applied to the benign images ($E_{d_B}$). Additionally, we determine the mean depth error for adversarial scenarios, involving the application of the defense technique to the adversarial image ($E_{d_c}$). Our findings indicate that our patch remains effective even when confronted with defense techniques (as depicted in Tables \ref{performance_1}, \ref{performance_2}, and \ref{performance_3}).

While the applied defense mechanisms do lead to a reduction in mean depth error, our patch continues to exhibit its influence, resulting in an average error of $0.2$. This corresponds to an approximate discrepancy of over $5$ meters in depth prediction. It's worth noting that the median blur defense significantly diminishes the baseline performance of the Monocular Depth Estimation (MDE) network.

\begin{table}[!htp]
  \centering
%%\small
  \caption{Mean depth error when applying JPEG compression.}
  \label{performance_1}
  \begin{tabular}{ccccc}
    \toprule
    \textbf{Parameters} &  \textbf{90} & \textbf{70} & \textbf{50}  & \textbf{30} \\
    \midrule
      $E_{d_c}$  & 0.23 & 0.22  & 0.19  & 0.15  \\
      $E_{d_B}$  &  0 & 0  & 0.001  & 0.001\\
  \bottomrule
\end{tabular}
\end{table}
 
\begin{table}[!htp]
  \centering
%\small
  \caption{Mean depth error when applying median blur.}
  \label{performance_2}
  \begin{tabular}{ccccc}
    \toprule
    \textbf{Parameters} &  \textbf{5} & \textbf{10} & \textbf{15}  & \textbf{20} \\
    \midrule
      $E_{d_c}$  & 0.13  & 0.12  & 0.13  & 0.15  \\
      $E_{d_B}$  & 0.09  &  0.08 & 0.06  & 0.08\\
  \bottomrule
\end{tabular}
\end{table}

\begin{table}[!htp]
  \centering
%\small
  \caption{Mean depth error when applying Gaussian noise.}
  \label{performance_3}
  \begin{tabular}{ccccc}
    \toprule
    \textbf{Parameters} &  \textbf{0.01} & \textbf{0.02} & \textbf{0.05}  & \textbf{0.1} \\
    \midrule
      $E_{d_c}$  & 0.24  & 0.21  &  0.2 & 0.21 \\
      $E_{d_B}$  &  0.01 & 0.03  & 0.05 & 0.089\\
  \bottomrule
\end{tabular}
\end{table}
\section{Related Work}\label{sec:rw}
In autonomous driving, prior research has predominantly centered around bolstering the security of perception sensors. This is evidenced by a multitude of attacks that have been devised with the aim of disrupting various components, including cameras \cite{Yan2016CanYT,Nassi_CCS, Petit_blackhat}, GPS systems \cite{Junjie_USENIX}, LiDAR systems \cite{Yulong_CCS, Hocheol_IACR}, RADAR sensors \cite{Radar}, ultrasonic sensors \cite{Yan2016CanYT}, and IMUs (Inertial Measurement Units) \cite{Yazhou_usenix, Trippel_EuroSP}. Other studies have concentrated on launching attacks tailored for regression tasks, such as optical flow estimation \cite{Ranjan_cvf} and depth estimation \cite{Yamanaka_Access}, while some have emphasized classification tasks, including 2D object detection and classification \cite{ykholt_usenix, Brown}, tracking \cite{Jia_ICLR}, traffic light detection \cite{Duan_cvpr}, and lane detection \cite{Sato}. In the context of our work, our primary focus revolves around disrupting Monocular Depth Estimation (MDE) models through jamming techniques.

Unlike existing physical attacks that have been directed at tasks like object detection \cite{Hu_ICCV, Thys_CVF}, image classification \cite{Athalye2018SynthesizingRA, ykholt_usenix}, and face recognition \cite{Sharif_CCS,Komkov_ICPR}, the area of attacks on depth estimation has received relatively limited attention. Zhang \cite{zhang_ECCV} proposes an attack technique designed to enhance performance in a universal attack scenario. Wong \cite{Wong_NIPS} introduces a strategy for generating targeted adversarial perturbations in images, which then randomly alter the associated depth map. It's worth noting that these two attacks center around digital perturbations, rendering them less suitable for real-world applications.

Yamanaka \cite{Yamanaka_Access} devises a method for generating printable adversarial patches, but the generated patch is trained to be applicable to random locations within the scene. On the other hand, Cheng \cite{Cheng_ECCV} concentrates on the inconspicuousness of the generated patch, ensuring that the patch remains unobtrusive and avoids drawing attention. However, the challenge with this patch lies in its object-specific nature, necessitating separate retraining for each target object. Furthermore, the patch's effectiveness is constrained by its limited affected region and its training for a specific context—namely, a fixed distance between the object and the camera—rendering it ineffective for varying distances.

Different from prior efforts, our emphasis lies in evaluating the comprehensive influence of the generated patch. We prioritize ensuring that the patch envelops the entirety of the target object, thus guaranteeing a thorough deception of the DNN-based vision system. Our framework introduces a methodology that ensures the efficacy of the patch across various objects within the same class, accommodating distinct shapes and sizes. Importantly, our patch is designed to function across varying distances between the object's placement and the camera of the vision system.
\section{Conclusion}\label{sec:conclusion}
Within this paper, we introduce a novel physical adversarial patch named APARATE, crafted with the explicit purpose of undermining Monocular Depth Estimation (MDE)-based vision systems. APARATE distinguishes itself as an adaptive adversarial patch, demonstrating the capacity to fully obscure objects or manipulate their perceived depth within a given scene, irrespective of their inherent size, shape, or placement.

Our empirical investigations provide compelling evidence for the effectiveness and resilience of our patch across diverse target objects. The achieved mean depth estimation error exceeds $14$ meters, with over $99\%$ of the target region undergoing alteration. Furthermore, our patch exhibits durability against defense techniques grounded in input transformations.

The repercussions of the proposed attack could lead to substantial harm, encompassing loss, destruction, and endangerment to both life and property. We assert that these findings should serve as a clarion call to the research community, stimulating the exploration of more robust and adaptive defense mechanisms.

%\begin{table}[htbp]
%\caption{Table Type Styles}
%\begin{center}
%\begin{tabular}{|c|c|c|c|}
%\hline
%\textbf{Table}&\multicolumn{3}{|c|}{\textbf{Table Column Head}} \\
%\cline{2-4} 
%\textbf{Head} & \textbf{\textit{Table column subhead}}& \textbf{\textit{Subhead}}& \textbf{\textit{Subhead}} \\
%\hline
%copy& More table copy$^{\mathrm{a}}$& &  \\
%\hline
%\multicolumn{4}{l}{$^{\mathrm{a}}$Sample of a Table footnote.}
%\end{tabular}
%\label{tab1}
%\end{center}
%\end{table}

%\begin{figure}[htbp]
%\centerline{\includegraphics{fig1.png}}
%\caption{Example of a figure caption.}
%\label{fig}
%\end{figure}

\section*{Acknowledgment}

%\section*{References}

\bibliographystyle{IEEEtran}
\bibliography{bib}
%\vspace{12pt}
%\color{red}
%IEEE conference templates contain guidance text for composing and formatting conference papers. Please ensure that all template text is removed from your conference paper prior to submission to the conference. Failure to remove the template text from your paper may result in your paper not being published.

\end{document}